\documentclass[conference]{IEEEtran}
\IEEEoverridecommandlockouts %
\usepackage{amsmath,amssymb,amsfonts}
\usepackage{amsthm, cite,enumitem}
\usepackage{url}
\usepackage{graphicx}

\usepackage{booktabs}
\usepackage{xcolor}
\newcommand{\rev}[1]{#1}

\graphicspath{{figs/}}
\usepackage{eso-pic}
\AddToShipoutPictureFG*{\AtPageUpperLeft{\put(\LenToUnit{0.5\paperwidth},\LenToUnit{-0.22in}){\makebox(0,0)[t]{\parbox{\textwidth}{\centering\scriptsize
Extended version of a paper accepted to IEEE MILCOM 2026 (Workshop S4), including the proofs omitted from the conference version.\\
\copyright~2026 IEEE. Personal use of this material is permitted. Permission from IEEE must be obtained for all other uses, in any current or future media, including reprinting/republishing this material for advertising or promotional purposes, creating new collective works, for resale or redistribution to servers or lists, or reuse of any copyrighted component of this work in other works.}}}}}

\theoremstyle{plain}
\newtheorem{proposition}{Proposition}
\newtheorem{corollary}{Corollary}
\newtheorem{remark}{Remark}

\begin{document}

\title{When Does Spatial Cooperation Help?\\Contamination Regimes in Decentralized\\Online Spectrum Anomaly Learning}

\author{\IEEEauthorblockN{Alp Berke Ardic and Erdem Koyuncu}
\IEEEauthorblockA{Department of Electrical and Computer Engineering,
University of Illinois Chicago\\
\{aardic2, ekoyuncu\}@uic.edu}
\thanks{This work was supported in part by the Army Research Lab under Grant W911NF-24-2-0172, and by the National Science Foundation under Grant 2531376.}
}

\maketitle

\begin{abstract}
Learning-based spectrum monitors adapt online to benign propagation changes, but persistent rogue transmissions can gradually become incorporated into the learned background. Spatially distributed receivers provide additional evidence for detecting such anomalies, while decentralized model mixing can also spread contamination to receivers with clean measurements. We study this tradeoff using a decentralized online learner with sample gating and model sharing. We show that, without defenses, contamination propagation is governed by the occupancy rate and is independent of cross-receiver correlation. In contrast, the benefit of spatial fusion depends strongly on correlation and the detector operating point, so cooperation can either improve or degrade detection. We also show that fragmented transmissions can evade gating based on temporally smoothed statistics. Combining smoothed and per-sample gating substantially improves robustness to such intermittent contamination.
\end{abstract}


\section{Introduction}
\label{sec:intro}

Electromagnetic spectrum operations increasingly rely on distributed,
learning-based monitors operating under continual distribution shift.
Propagation conditions drift, legitimate occupancy changes, and hardware
responses wander with temperature. A common solution is online or
test-time adaptation, in which a deployed model keeps learning from the
measurements it receives \cite{sun2025radiopit,li2026robust}. In a
contested spectrum, adaptation carries a corresponding risk. When a
persistent unauthorized transmitter appears and its emissions are
repeatedly admitted into the adaptation set, the model gradually accepts
the intruder as part of the background and loses the ability to raise an alarm while the rogue remains present \cite{schosser2025advancing}.
\rev{Persistent contamination can therefore cause an adaptive detector to fail to detect the anomaly even while the underlying rogue signal remains present.}

Spatially distributed receivers offer a possible solution, since several
receivers observe the same physical event through different channels and
pooled evidence sharpens spectrum situational awareness.
Distributed and federated architectures for spectrum sensing and radio map
estimation are well developed \cite{shi2022federated,zhang2024fedrme,
ma2025fedrme}, and decentralized learners that mix
model parameters with neighbors form a classical line of work
\cite{nedic2009distributed,xiao2004fast}, alongside edge protocols for walk-based training \cite{ardic2023random} and model-distributed inference \cite{ardic2025decentralized}.
However, cooperation carries a risk of its own. 
Several receivers can be contaminated at once, and model mixing opens a second contamination pathway that no amount of local vigilance closes, as a receiver that never hears the rogue can still inherit a poisoned model from a neighbor. Therefore, we do not assume adding receivers reduces contamination, and instead ask under which contamination conditions cooperation helps.

\begin{figure}
\centering
\includegraphics[width=\columnwidth]{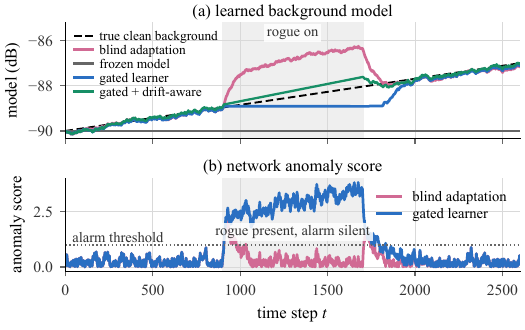}\vspace{-10pt}
\caption{An $800$-sample rogue dwell, i.e., a continuous interval during which the rogue transmitter remains active. The blindly adapting learner gradually incorporates the rogue into its background model, causing the anomaly score to fall below the alarm threshold while the rogue remains present. Gating prevents this adaptation and keeps the alarm active, but the frozen background estimate becomes stale as the environment drifts. The drift-aware variant continues to track the background while the gate is closed.}
\label{fig:demo}
\end{figure}

Fig.~\ref{fig:demo} illustrates the basic difficulty created by online adaptation. A blindly adapting learner can gradually absorb a persistent rogue into its background model and stop raising an alarm, while gating prevents this contamination but can leave the learned background stale as the environment drifts. These competing effects motivate the contamination, detection, and adaptation mechanisms studied in the rest of the paper.

We consider a decentralized online learning setting that couples sample gating with model mixing among neighboring receivers. Contamination is described by two operational parameters: the fraction of time each receiver is affected and the tendency of receivers to be affected together. The latter separates a single strong emitter heard everywhere from independent local interferers. The contributions of this paper are then summarized below.
\begin{itemize}[leftmargin=*,labelsep=0.4em]
\item We formulate contamination-aware decentralized online RF learning in a form that separates physical contamination of measurements from algorithmic contamination reaching clean receivers through model mixing, under a process governed by an occupancy rate and a cross-receiver
correlation\rev{.}
\item We prove that decentralized learning  without defenses propagates contamination according to the marginal occupancy rate and is independent of cross-receiver correlation. Cross-receiver correlation affects performance only when defenses exploit the joint contamination pattern across receivers\rev{.}
\item We characterize mean-fusion detection and identify a computable boundary separating the regimes in which correlated contamination improves or degrades detection performance, with detection probabilities available in closed form.
\item We show that gating based only on a smoothed statistic can miss fragmented/intermittent transmissions. Combining the smoothed test with a per-sample test substantially improves detection of such transmissions with negligible extra cost.
\end{itemize}

The rest of this paper is organized as follows. Section~\ref{sec:related} reviews related work. Section~\ref{sec:model} introduces the system, contamination, and detection models. Section~\ref{sec:theory} develops the theoretical results on contamination propagation and mean-fusion detection. Section~\ref{sec:results} presents the numerical results, including fragmentation, background drift, and a larger network. Section~\ref{sec:design} summarizes the resulting design implications, and Section~\ref{sec:conclusion} concludes the paper. 

\section{Related Work}
\label{sec:related}

\textbf{Radio map estimation and adaptation.} Deep radio map estimators
such as RadioUNet \cite{levie2021radiounet} are typically trained offline.
RadioPiT \cite{sun2025radiopit} adds test-time self-supervision to adapt
a pretrained model in the field, and the online scheme of
\cite{li2026robust} raises its learning rate on novel samples. Nonetheless, this policy would absorb a persistent anomaly faster. 
Selective test-time adaptation curates which samples may update the model \cite{ambekar2024selective,kim2026candi}, in single-model centralized settings. 
Federated spectrum sensing \cite{shi2022federated} and radio map estimation \cite{zhang2024fedrme,ma2025fedrme,oral2026repa} distribute the computation while assuming clean adaptation streams.

\textbf{Cooperative sensing under attack.} Cooperative sensing has been
studied under spectrum-sensing data falsification
\cite{rawat2011collaborative} and primary-user emulation
\cite{chen2008defense}, and correlated shadowing is known to limit the
diversity gain of cooperation \cite{ghasemi2005collaborative}. In the
falsification literature malicious sensors misreport a fixed detection
task, while here truthful receivers suffer physically contaminated
measurements and the null hypothesis itself is learned online as
contamination drags it toward the anomaly. \rev{Correlated shadowing captures the familiar loss of diversity under correlation. In the adaptive setting studied here, however, the effect of correlation depends on the detector operating point and need not have a universal ordering.}

\textbf{Robust learning and evasion.} Byzantine-robust aggregation \cite{blanchard2017machine} defends
parameter averaging against adversarial updates. Contamination here enters through the measurements of legitimate receivers.
Rate and correlation grade its severity rather than a count of corrupted workers.
Classical distributed detection \cite{varshney1997distributed} fuses static hypotheses. Evasion through mismatched timescales appears in low-rate attacks on TCP
\cite{kuzmanovic2003low} and gradual poisoning of anomaly detectors \cite{rubinstein2009antidote}. 
Prior work treats adaptation hygiene, distributed sensing, and robust aggregation separately, and this paper characterizes their interaction under one correlated contamination process.

\section{System Model}
\label{sec:model}

\subsection{Observations and Learner}

\begingroup
We consider $N$ receivers indexed by $\mathcal{V}=\{0,\ldots,N-1\}$ that form a connected graph. Receiver $i$ maintains a scalar background-power model $\theta_i(t)$ in dB and observes
\begin{equation}
\label{eq:obs}
x_i(t)=b(t)+a_iC_i(t)+n_i(t).
\end{equation}
Here $b(t)$ is a slowly varying clean background. The quantity $a_i\ge0$ is the received rogue power. The indicator $C_i(t)\in\{0,1\}$ specifies whether receiver $i$ is contaminated. The term $n_i(t)\sim\mathcal{N}(0,\sigma^2)$ is independent noise. We write $\mathbf a=(a_0,\ldots,a_{N-1})$ and $\mathbf C(t)=(C_0(t),\ldots,C_{N-1}(t))$. If $a_i=0$, receiver $i$ is physically clean and any contamination-induced model bias must arrive through cooperation.

We use a standard consensus-plus-local-gradient decentralized update~\cite{nedic2009distributed}, specialized here to quadratic loss:
\begin{equation}
\label{eq:update}
\theta_i(t{+}1)=\textstyle \sum_{j\in\mathcal N_i}\widetilde W_{ij}(t)\theta_j(t)
+\eta q_i(t)\bigl(x_i(t)-\theta_i(t)\bigr).
\end{equation}
Here $W$ is a doubly stochastic mixing matrix consistent with the graph. The neighborhood $\mathcal N_i$ includes receiver $i$. The learning rate is $\eta>0$. Variable $q_i(t)\in\{0,1\}$ gates the local sample. The matrix $\widetilde W(t)$ is the effective mixing matrix after excluding any neighbors that are deemed untrustworthy by the defense mechanism. When no neighbors are excluded, $\widetilde W=I+\gamma(W-I)$ with $0<\gamma\le1$. The parameter $\gamma$ controls the strength of inter-receiver mixing: Smaller $\gamma$ corresponds to slower information exchange relative to local learning.
\endgroup

\textbf{Threat model.} \rev{The rogue may control its transmission timing and received-power trajectory, including fragmented bursts or gradual ramps, but cannot alter links, model updates, or receiver reports. Thus contamination enters only through the physical measurements of legitimate receivers.}

\subsection{Contamination Process}

\begingroup
Let $p\in[0,1]$ denote the marginal contamination probability, or occupancy rate, at each receiver. Thus, $\mathbb{E}[C_i(t)]=p$. Let $\rho_C\in[0,1]$ denote the pairwise cross-receiver correlation of the contamination indicators. To vary $\rho_C$ without changing $p$, let $S(t)$ be a shared schedule and $I_i(t)$ independent receiver-specific schedules with the same marginal occupancy and temporal law. With independent $R_i\sim\mathrm{Bernoulli}(\sqrt{\rho_C})$, set
$C_i(t)=R_iS(t)+(1-R_i)I_i(t)$.
Thus, each receiver follows either the shared contamination schedule $S(t)$
or its receiver-specific schedule $I_i(t)$. Averaging over the random
choices $R_i$, every receiver has occupancy rate $p$ and every receiver
pair has correlation $\rho_C$. This construction allows arbitrary temporal
dependence in the contamination process. In the numerical experiments, we use blockwise schedules as particular instances of this general model, since they allow us to vary the duration and fragmentation of contamination in a controlled manner.
\endgroup

\subsection{Scores, Gates, and Detectors}

\begingroup
Each receiver tracks an exponential moving average $\zeta_i(t)$ of the standardized innovation $(x_i-\theta_i)/\sigma$ at rate $\alpha\in(0,1]$, corresponding to a memory of order $1/\alpha$ samples, and shares this scalar with its neighbors. For $\tau_g>0$, local gating uses
\begin{equation}
\label{eq:localgate}
q_i^{\mathrm{loc}}(t)=\mathbf 1\bigl\{|\zeta_i(t)|\le\tau_g\bigr\}.
\end{equation}
With $\zeta_i^f(t)=|\mathcal N_i|^{-1/2}\sum_{j\in\mathcal N_i}\zeta_j(t)$, spatial gating uses
\begin{equation}
\label{eq:spatialgate}
q_i^{\mathrm{sp}}(t)=\mathbf 1\bigl\{|\zeta_i^f(t)|\le\tau_g
\ \text{or}\ |\zeta_i(t)|\le\lambda\tau_g\bigr\},
\end{equation}
where $0<\lambda\le1$ prevents fused evidence alone from freezing a receiver whose own data look strongly clean. The proposed learner also excludes neighbor $j$ from model mixing if $j$ reports an anomalous state or if an exponential average of $(x_i-\theta_j)/\sigma$, updated at rate $\alpha_T$, exceeds the trust threshold $\tau_T$ in magnitude. The corresponding mixing weight is reassigned to receiver $i$'s diagonal weight. Under ideal clean data, $\mathrm{std}(\zeta_i)=\sqrt{\alpha/(2-\alpha)}$.

Throughout, an \emph{alarm} is raised whenever the rogue is detected by an individual receiver or by a fused detection rule. For detection analysis, let $z_i=(x_i-b)/\sigma$. Under the clean hypothesis, $z_i\sim\mathcal{N}(0,1)$. At target one-sided false-alarm probability $\delta$, the single-receiver and mean-fusion tests use $\tau=Q^{-1}(\delta)$. For the max-fusion test, the threshold is $\tau_{\max}=\Phi^{-1}((1-\delta)^{1/N})$. Thus, all three rules are matched to the same false-alarm level without fixing any numerical operating point in the system model.

\endgroup

\section{Theoretical Properties}
\label{sec:theory}

In this section, we establish theoretical properties of contamination propagation under decentralized learning and characterize the effect of mean fusion on anomaly detection.

\subsection{Invariance of Contamination Propagation}

We first ask how contamination travels through the mixing step when no
defense is active. The answer takes a simple form because the ungated
update is linear.

For compact notation, define the network model-state vector as $\boldsymbol{\theta}(t)=(\theta_0(t),\ldots,\theta_{N-1}(t))^\top$ and the observation vector as $\mathbf{x}(t)=(x_0(t),\ldots,x_{N-1}(t))^\top$.
\begin{proposition}
\label{prop:linear}
Consider unprotected decentralized learning, that is, $q_i \equiv 1$ with a
fixed doubly stochastic $W$. The expected trajectory
$\mathbb{E}[\boldsymbol{\theta}(t)]$ of the network model-state vector depends on the contamination process
only through the per-receiver means $s \mapsto \mathbb{E}[\mathbf{C}(s)]$.
\end{proposition}
\begin{proof}
With $q \equiv 1$ the update \eqref{eq:update} is linear and time
invariant. Writing $M = \widetilde{W} - \eta I$ gives
$\boldsymbol{\theta}(t{+}1) = M\boldsymbol{\theta}(t) +
\eta\,\mathbf{x}(t)$, and by unrolling,
$\boldsymbol{\theta}(t) = M^t \boldsymbol{\theta}(0) + \eta \sum_{s<t}
M^{t-1-s}\mathbf{x}(s)$. Taking expectations, we have
$\mathbb{E}[\mathbf{x}(s)] = b(s)\mathbf{1} + \mathbf{a} \odot
\mathbb{E}[\mathbf{C}(s)]$. The joint distribution of the indicators
across receivers never enters.
\end{proof}

The expected trajectory of the network model-state vector is therefore unaffected by the cross-receiver correlation $\rho_C$ and by the temporal arrangement of contamination whenever the per-receiver contamination means remain constant over the contamination window.

\begin{corollary}
\label{cor:affine}
Let Receiver $j$ be physically clean, $a_j = 0$, and connected to
contaminated receivers. Its steady-state expected bias equals
$\kappa\, p + \beta$, where $\kappa$ depends only on
$(\mathbf{a}, \widetilde{W}, \eta)$ and $\beta$ is the drift lag present
with no contamination at all.
\end{corollary}
\begin{proof}
Under constant marginal occupancy $\mathbb{E}[\mathbf C(t)]=p\mathbf 1$, Proposition~\ref{prop:linear} implies that the contamination-dependent part of $\mathbb{E}[\boldsymbol{\theta}(t)]$ is linear in $p$. Whenever the expected learner dynamics reach a steady state, the bias at receiver $j$ therefore has the form $\kappa p+\beta$, where $\beta$ is the bias obtained at $p=0$ and $\kappa$ depends on $\mathbf a$, $\widetilde W$, and $\eta$.
\end{proof}

\begin{remark}
\label{rem:defense}
Any dependence of contamination propagation on $\rho_C$ must arise through sample gating or data-dependent model mixing. Under unprotected mixing with fixed receiver placement, the expected model error at a physically clean receiver grows linearly with the rogue's duty cycle, or occupancy rate $p$. A shared rogue therefore produces the same expected model error as independent interferers that affect the same receivers with the same occupancy rate, even though their cross-receiver correlations are very different. The coefficient $\kappa$ depends on which receivers are directly affected by the rogue and on their received rogue powers.
\end{remark}

\subsection{Fusion Regimes Under Correlated Contamination}
We next characterize how cross-receiver correlation affects detection based on mean fusion. Let $D=(\sigma\sqrt{N})^{-1}\sum_i a_i C_i$ denote the normalized mean shift of the mean-fusion statistic. Since the normalized statistic has unit variance under the clean hypothesis, $D$ measures the strength of the aggregate anomaly relative to the noise level. For a fixed alarm threshold, a larger $D$ corresponds to a higher probability of detection. Let $\Phi$ denote the standard normal distribution function and let $Q(x)=1-\Phi(x)$ denote its tail probability.

\begin{proposition}
\label{prop:fusion}
Fix a one-sided false-alarm probability $\delta$ and let $\tau=Q^{-1}(\delta)$. Under the contamination model above, the following properties hold.
\begin{enumerate}
\item[(i)] \emph{Single-receiver detection.}
Receiver $i$ raises an alarm with probability
$P_{D,i}^{\rm single}=p\,Q(\tau-a_i/\sigma)+(1-p)Q(\tau)$. This depends only on the marginal occupancy rate $p$, and is independent of the cross-receiver correlation $\rho_C$.

\item[(ii)] \emph{Mean-fusion scaling.}
If only receiver $i$ is contaminated, then $D=a_i/(\sigma\sqrt{N})$, which is $1/\sqrt{N}$ times the normalized anomaly strength obtained by testing receiver $i$ alone. Thus, averaging with clean receivers dilutes an anomaly that is present at only one receiver. If all $N$ receivers are contaminated with the same amplitude $a$, then $D=a\sqrt{N}/\sigma$. In this case, mean fusion increases the normalized anomaly strength by a factor of $\sqrt{N}$ relative to a single receiver.

\item[(iii)] \emph{Effect of cross-receiver correlation.}
Let $D_{\max}=\sum_i a_i/(\sigma\sqrt{N})$ denote the largest possible normalized mean shift, corresponding to simultaneous contamination of all receivers. Under the contamination model of Section~\ref{sec:model}, if $D_{\max}\le\tau$, then $P_D=\mathbb{E}[\Phi(D-\tau)]$ is non-decreasing in $\rho_C$. Thus, increasing cross-receiver correlation cannot reduce the mean-fusion detection probability.
\end{enumerate}
\end{proposition}
\begin{proof}
For (i), condition on $C_i$. The decision rule uses only receiver $i$, so no cross-receiver quantity appears and the stated mixture follows. Item (ii) follows by evaluating $D$ when only one receiver is contaminated and when all receivers are contaminated. For (iii), first note that $\mathbb{E}[D]=p\sum_i a_i/(\sigma\sqrt{N})$, which is independent of $\rho_C$.

To compare different values of $\rho_C$, let $U_i\sim\mathrm{Uniform}(0,1)$ be fixed for each receiver and set
$R_i(\rho_C)=\mathbf{1}\{U_i\le\sqrt{\rho_C}\}$.
As $\rho_C$ increases, some receivers switch from their independent contamination schedule $I_i(t)$ to the shared schedule $S(t)$. This switch does not change the marginal contamination probability of any receiver, but it increases the dependence among receivers that follow the shared schedule.

For random variables with fixed marginals, making them more positively dependent increases their weighted sum in the convex order \cite{shaked2007stochastic}. Therefore, each such switch increases the normalized mean shift $D$ in the convex order while leaving $\mathbb{E}[D]$ unchanged. Repeating this argument over the receivers shows that $D$ increases in the convex order as $\rho_C$ increases. Hence, the normalized mean shift $D$ increases in the convex order as $\rho_C$ increases. Moreover, every attainable value of $D$ lies in $[0,D_{\max}]$. When $D_{\max}\le\tau$, the map $u\mapsto\Phi(u-\tau)$ is convex over this entire interval. Thus, $\mathbb{E}[\Phi(D-\tau)]$, and hence $P_D$, is non-decreasing in $\rho_C$.
\end{proof}

\begin{remark}
	\label{rem:regimes}
	By Item~(iii), when $D_{\max}\le\tau$, increasing cross-receiver correlation cannot reduce the mean-fusion detection probability. This regime corresponds to relatively weak anomalies or stringent false-alarm requirements. When $D$ exceeds the alarm threshold, this monotonicity is no longer guaranteed, and increasing correlation can instead reduce the fusion gain, as shown by the operating points in Table~\ref{tab:regime}. %
\end{remark}

\section{Numerical Results}
\label{sec:results}

\begingroup
Unless otherwise stated, the main experiments use a four-receiver chain with $\mathbf a=(1.8,1.8,0.8,0)$\,dB, $\sigma=1$, $\gamma=0.05$, $\eta=0.015$, $\alpha=0.1$, $\lambda=1/2$, $\alpha_T=0.02$, $\tau_T=0.5$, $\tau_g=1$, and $P_{FA}=10^{-2}$. Receiver $3$ is physically clean. The learner and detector memories are about $67$ and $10$ samples, respectively. The clean-score standard deviation is about $0.23$. Occupancy--correlation experiments use $40$-sample blocks, $T=2600$, a $400$-sample contamination window, and a $+3$\,dB background drift. Active-block counts are matched so each realized marginal occupancy equals $p$. Learner metrics use $20$ seeds per $(p,\rho_C)$ cell and detection metrics use $200$. A separate experiment uses a nine-receiver grid.
\endgroup

\begin{table}[t]
\caption{Gain of mean fusion over the best single receiver at $p = 0.5$,
in percentage points of detection probability, computed by exact exhaustive enumeration.}
\label{tab:regime}
\centering
\begin{tabular}{lccc}
\toprule
Operating point & $\rho_C{=}0$ & $\rho_C{=}0.5$ & $\rho_C{=}1$ \\
\midrule
$P_{FA}{=}1\%$, $\mathbf{a}$& $+0.1$ & $+3.8$ & $+7.5$ \\
$P_{FA}{=}20\%$, $\mathbf{a}$ & $+6.9$ & $+5.5$ & $+4.1$ \\
$P_{FA}{=}1\%$, $3\times\mathbf{a}$ & $+16.0$ & $+8.0$ & $+0.1$ \\
$P_{FA}{=}1\%$, $4\times\mathbf{a}$ & $+25.0$ & $+12.5$ & $+0.0$ \\
\bottomrule
\end{tabular}
\end{table}

\begin{figure*}[t]
\centering
\includegraphics[width=\textwidth]{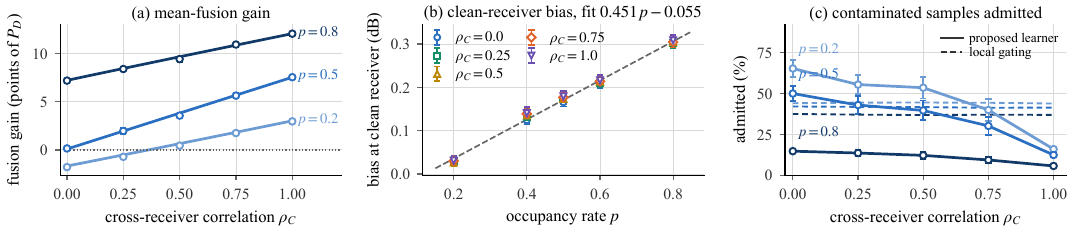}\vspace{-8pt}
\caption{Behavior over the occupancy rate and correlation plane. (a)~Mean-fusion
gain over the strongest single receiver. Lines show exact theory and
points show simulation with $200$ seeds and $\pm 1$ SEM bars. (b)~Bias
propagated into the clean receiver under unprotected mixing. All five
correlation levels land on one line in the rate, as
Proposition~\ref{prop:linear} requires. (c)~Contaminated samples
admitted. The proposed learner, solid, exceeds local gating, dashed, at
low rates when $\rho_C = 0$ and crosses below it as $p$ and $\rho_C$
grow. Panels (b) and (c) use $20$ seeds per cell.}
\label{fig:grid}
\end{figure*}

\subsection{Exact Theory Against Simulation}
Fig.~\ref{fig:grid}a overlays the closed-form mean-fusion gain on Monte Carlo estimates across three occupancy rates and five correlation levels. Agreement is within $0.34$ points of $P_D$ everywhere. The Monte Carlo run shares the residual model with the enumeration, so it cross-checks implementations rather than modeling assumptions. The gain curves rise with $\rho_C$ throughout, as Proposition~\ref{prop:fusion}(iii) requires at this operating point, where the largest normalized mean shift is $D_{\max}=2.20$, compared with $\tau=2.33$. Table~\ref{tab:regime} shows how this behavior changes at other operating points. When $D_{\max}\le\tau$, the mean-fusion gain increases with cross-receiver correlation. When this condition is violated, the trend can reverse and the gain can decrease as correlation increases. At $p = 0.2$ and $\rho_C \lesssim 0.4$, the gain is negative, since fusion with mostly clean neighbors dilutes a real local anomaly below what the receiver sees alone. Max fusion, omitted from the plot, gains at most $1.2$ points anywhere and shrinks with correlation, since the OR rule's threshold penalty consumes the coverage it adds.

\subsection{Propagation Into Clean Receivers}

Fig.~\ref{fig:grid}b shows the late-window bias of the physically clean
receiver under unprotected mixing. Across all $25$ cells the five
correlation levels are statistically indistinguishable, spread $0.006$
to $0.017$\,dB at fixed occupancy rate against a per-cell standard error of
$0.011$\,dB, and collapse onto $0.451\,p - 0.055$\,dB with
$R^2 = 0.995$. The intercept is a prediction of
Corollary~\ref{cor:affine}, since it should equal the pure drift lag, and
an independent contamination-free measurement gives
$-0.060 \pm 0.012$\,dB.

\subsection{Screening Over the Occupancy Rate and Correlation}

Rate and correlation together decide whether spatial cooperation pays. Fig.~\ref{fig:grid}c shows the fraction of contaminated samples admitted into adaptation. 
Local gating is flat at $37\%$ to $44\%$. This is expected, since each receiver's own score sees the joint pattern only weakly, through the mixed models. 
The proposed learner starts worse. At $\rho_C = 0$ it admits $65\%$ against $44\%$ for local gating at $p = 0.2$, and $50\%$ against $42\%$ at $p = 0.5$. 
As correlation grows it crosses below, and at $\rho_C = 1$ it admits $12\%$ at $p = 0.5$, a roughly $3.3\times$ advantage. 
A decomposition at $p = 0.5$ traces the low-correlation excess to neighbor exclusion rather than the fused gate.
Spatial gating with fixed mixing admits $41.4 \pm 3.7\%$, which is indistinguishable from local gating. The neighbor-exclusion mechanism raises admissions by a paired $8.6 \pm 2.2$ points. When a receiver excludes clean neighbors from model mixing, it loses their stabilizing influence, absorbs its local anomaly more strongly, and its gate reopens sooner.
The surface also resists any one-parameter summary. 
The cells $(0.2, 1)$ and $(0.5, 0.25)$ in $(p, \rho_C)$ carry nearly matched probabilities of simultaneous contamination, $0.20$ against $0.219$. Yet they admit $15.9\%$ against $42.9\%$.

\begin{figure*}[t]
\centering
\includegraphics[width=\textwidth]{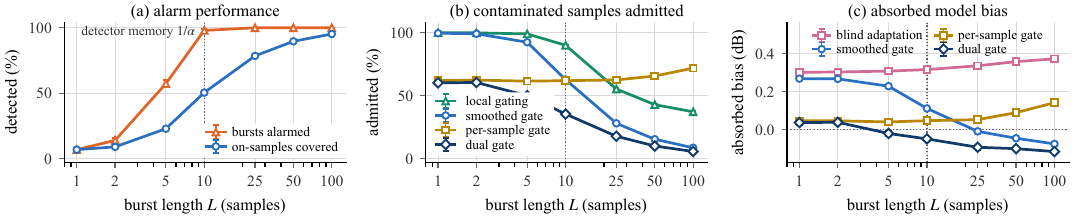}\vspace{-8pt}
\caption{Fragmentation of a fixed transmit budget, $100$ contaminated
samples of a $400$-sample window with all reachable receivers hit
together, $20$ seeds, $\pm 1$ SEM. Only the burst length $L$ varies, and
the per-sample clean-reference detection rate stays at $44$ to $46\%$.
(a)~Alarm performance collapses below the detector memory
$1/\alpha = 10$, counted per burst (upper) and per contaminated sample
(lower). (b)~Smoothed gates admit nearly everything at small $L$, the
per-sample gate is timing invariant, and the dual gate admits the fewest
at every $L$. (c)~The blind learner's absorbed bias barely moves, while
the dual gate never absorbs more than $+0.04$\,dB, at the price of a
small stale-model lag at long bursts.}
\label{fig:frag}
\end{figure*}

\subsection{Intermittent Contamination and Detector Timescales}
The preceding results concern whether contamination is shared across receivers. The experiment in Fig.~\ref{fig:frag} instead examines the temporal structure of contamination. The detector averages evidence over an effective timescale of about $1/\alpha=10$ samples, whereas the learning update averages admitted samples over the longer timescale $1/\eta\approx67$ samples. The rogue has a nominal budget of $100$ contaminated samples, which it distributes into bursts of length $L$. The contamination amplitude at each affected sample is unchanged as $L$ varies. When $L$ is shorter than the detector's averaging timescale, however, the smoothed statistic does not accumulate enough evidence to cross the alarm threshold. Consequently, alarm coverage falls from $95.2\%$ at $L=100$ to $6.7\%$ at $L=1$ (Fig.~\ref{fig:frag}a), while the fraction of contaminated samples admitted by the smoothed gate rises from $8.6\%$ to $99.6\%$ (Fig.~\ref{fig:frag}b). The resulting absorbed bias increases from $-0.07$\,dB  to $+0.27$\,dB (Fig.~\ref{fig:frag}c), reaching within $11\%$ of blind adaptation. The transition occurs near $L\approx1/\alpha$, consistent with the detector's effective averaging timescale.

In contrast, the blind learner changes only from $+0.30$ to $+0.37$\,dB across the same hundredfold variation in $L$. This weak dependence on burst length is consistent with Proposition~\ref{prop:linear}, which shows that the expected unprotected learning trajectory is insensitive to the temporal arrangement of contamination when the per-time contamination means are unchanged. The small remaining variation is attributable to the finite contamination window. Bursts that overlap its boundaries are partially truncated, causing small deviations in $\mathbb{E}[C_i(t)]$ near the beginning and end of the window.

The poor performance for short bursts is caused by relying on a single smoothed gating statistic. To address this limitation, Figs.~\ref{fig:frag}b and \ref{fig:frag}c also evaluate a per-sample gate that applies the two-evidence structure of \eqref{eq:spatialgate} directly to the raw innovation. Because it does not average evidence over time, its response is largely insensitive to burst length. It admits about $62\%$ of contaminated samples (Fig.~\ref{fig:frag}b) and, for $L\le25$, keeps the absorbed bias near $+0.05$\,dB (Fig.~\ref{fig:frag}c). The corresponding clean-data rejection rate is only $1.7\%$.

For long bursts, however, the per-sample gate becomes less effective because many contaminated samples remain below its instantaneous threshold. At $L=100$, it admits $71.8\%$ of contaminated samples and yields an absorbed bias of $+0.14$\,dB, whereas the smoothed gate admits only $8.6\%$ and yields $-0.07$\,dB. The two gates are therefore effective in complementary regimes. Their conjunction, which we call the dual gate, admits the fewest contaminated samples over the full range of $L$ (Fig.~\ref{fig:frag}b). These results show that combining smoothed and per-sample gating reduces the effectiveness of fragmentation attacks but does not eliminate them completely.

\subsection{Adaptation Under Background Drift}

We now return to the example in Fig.~\ref{fig:demo}. In this experiment, the rogue remains continuously active for an $800$-sample dwell, compared with the $400$-sample contamination windows used in the preceding occupancy--correlation experiments. The blindly adapting learner gradually incorporates the rogue into its background model. Its anomaly score falls below the alarm threshold roughly $150$ samples after the dwell begins, even though the rogue remains present. The gated learner prevents contaminated samples from entering the model and therefore keeps the alarm active throughout the dwell. However, because the background model is frozen while the true background continues to drift, its estimate becomes increasingly stale. After the rogue disappears, this mismatch slows recovery. Over $20$ seeds, the frozen-gating scheme has a late-dwell bias of $-0.79$\,dB and requires $306\pm12$ samples to recover.

We therefore consider a drift-aware variant that estimates both the background level and its rate of change. While the gate is closed, the background estimate continues to evolve according to the estimated drift rather than being frozen. This reduces the late-dwell bias to $+0.36\pm0.05$\,dB and shortens recovery to $50\pm8$ samples. The improvement, however, introduces a vulnerability to gradually increasing contamination. In a ramp experiment, the rogue increases its received power to $1.8$\,dB over $1200$ samples. The gradual increase remains below the gating thresholds in all seeds, causing every learner to incorporate the contamination into its background model. The drift-aware learner also extrapolates the contaminated trend, reaching $1.84$\,dB compared with $1.73\pm0.01$\,dB for frozen gating and overshooting to $2.01$\,dB after the ramp levels off. Thus, drift-aware adaptation can substantially improve recovery after gating, but the estimated rate of change must itself be protected against gradual contamination.

\subsection{A Larger Network}

We repeated the core comparisons over ten seeds on a $3 \times 3$ grid
of nine receivers, the rogue near one corner and the opposite corner
physically clean. The same pattern appears, sharper because the larger $\sqrt{N}$
dilutes single-receiver evidence more. The mean-fusion gain runs $-3.6$,
$+0.8$, $+4.9$ points at $\rho_C = 0$, $0.5$, $1$ for $p = 0.5$. Bias at
the clean corner stays flat at $+0.041$ to $+0.047 \pm 0.012$\,dB. Local
gating admits $80$ to $81\%$ at every correlation, while the proposed
learner starts at $83.4\%$ and drops to $52.8\%$.

\section{Implications for System Design}
\label{sec:design}

\textbf{Cooperation regimes.} The dangerous regime occurs when the cross-receiver contamination correlation $\rho_C$ is low and the occupancy rate is low or moderate. The propagation cost of mixing is paid in full there. Spatial detection and gating provide little benefit in this regime, while excluding neighbors deemed untrustworthy may occur too late to prevent contamination from propagating (Sec.~\ref{sec:results}-C). Uncorrelated interference favors local gating with weak mixing. 

\textbf{Detector design.} The fragmentation experiment shows that smoothed gating alone is vulnerable to short bursts. Combining smoothed and per-sample tests provides protection against both short and persistent contamination, while sequential accumulation \cite{page1954continuous} provides another way to retain evidence over time. Burst fragmentation at fixed energy mirrors low-probability-of-intercept emission practice, so the dual gate addresses a tactic already in the field.

\textbf{Adaptation during gating.} Freezing the learner while its gate is closed prevents contaminated samples from entering the model but can leave the background estimate stale when the environment continues to drift. The numerical results show that drift-aware adaptation can accelerate recovery, but gradual contamination can corrupt the estimated drift. Thus, adaptation during gating should account for background drift while protecting both the model level and its rate of change.

\section{Conclusion}
\label{sec:conclusion}

This paper studied when spatial cooperation helps decentralized online spectrum anomaly learning and when it propagates contamination. Under unprotected mixing, the expected contamination transferred to clean receivers depends on the occupancy rate but not on cross-receiver correlation. In contrast, the benefit of fusion depends on both correlation and the operating point, with a computable boundary separating favorable and unfavorable regimes. The numerical results further demonstrate the importance of contamination timing and background drift for robust online learners. Extending these results to richer models and to hardware validation is left for future work.

\bibliographystyle{IEEEtran}
\bibliography{refs}

\end{document}